\documentclass[10pt,twocolumn]{article}

\usepackage[utf8]{inputenc}
\usepackage{amsmath}
\usepackage{amssymb}
\usepackage{titlesec}
\usepackage{cite}
\usepackage{booktabs}
\usepackage{graphicx}
\usepackage[colorlinks=false]{hyperref}
\usepackage[format=plain,labelfont=it]{caption}
\usepackage[left=1.5cm,right=1.5cm,top=2cm,bottom=2cm]{geometry}
\usepackage{xcolor}
\usepackage{stmaryrd}
\usepackage{multirow,array}
 \usepackage{mathtools,cuted}
 \usepackage{subfig}
\titleformat{\section}{\centering\normalfont\scshape}{\Roman{section}.}{5pt}{}
\titleformat{\subsection}{\normalfont\it}{\Alph{subsection}.}{5pt}{}
\titleformat{\subsubsection}{\normalfont\it}{\hspace{4mm}\arabic{subsubsection})}{5pt}{}

\newcommand\infoFootnote[1]{%
  \begingroup
  \renewcommand\thefootnote{}\footnote{#1}%
  \addtocounter{footnote}{-1}%
  \endgroup}

\newcommand{\preprintNoteIEEEAccepted}[1]{%
    © {#1} IEEE.  Personal use of this material is permitted.  Permission from IEEE must be obtained for all other uses, in any current or future media, including reprinting/republishing this material for advertising or promotional purposes, creating new collective works, for resale or redistribution to servers or lists, or reuse of any copyrighted component of this work in other works.}

\newcommand{\uproman}[1]{\uppercase\expandafter{\romannumeral#1}}

\newcommand{\R}{\mathbb{R}}

\newcommand{\yb}{\boldsymbol{y}}

\newcommand{\xb}{\boldsymbol{x}}

\newcommand{\wb}{\boldsymbol{w}}

\newcommand{\mb}{\boldsymbol{m}}

\newcommand{\wbt}{\tilde{\boldsymbol{w}}}

\newcommand{\sk}{\mathtt{sk}}
\newcommand{\pk}{\mathtt{pk}}
\newcommand{\Enc}{\mathsf{Enc}}

\newcommand{\Dec}{\mathsf{Dec}}

\newcommand{\cipher}[1]{\llbracket {#1}\rrbracket}

\newcommand{\blue}[1]{#1}
\newcommand{\orange}[1]{#1}

\DeclareMathAlphabet\mathbfcal{OMS}{cmsy}{b}{n}

\def\tvdots{\vbox{\baselineskip=2pt \lineskiplimit=0pt \kern6pt \Gbox{.}\Gbox{.}\Gbox{.}}} 
\title{\bf Privacy-preserving gradient-based fair federated learning}

\author{Janis Adamek, and Moritz Schulze Darup}% 

\begin{document}
\maketitle
\thispagestyle{empty}
\pagestyle{empty}
	
%%%%%%%%%%%%%%%%%%%%%%%%%%%%%%%%%%%%%%%%%%%%%%%%%%%%%%%%%%%%%%%%%%%%%%
\maketitle

\textbf{\textit{Abstract}.} {\bf Federated learning (FL) schemes allow multiple participants to collaboratively train neural networks without the need to directly share the underlying data. However, in early schemes, all participants eventually obtain the same model. Moreover, the aggregation is typically carried out by a third party, who obtains combined gradients or weights, which may reveal the model. These downsides underscore the demand for fair and privacy-preserving FL schemes. Here, collaborative fairness asks for individual model quality depending on the individual data contribution. Privacy is demanded with respect to any kind of data outsourced to the third party. Now, there already exist some approaches aiming for either fair or privacy-preserving FL and a few works even address both features.
In our paper, we build upon these seminal works and present a novel, fair \text{and} privacy-preserving FL scheme. Our approach, which mainly relies on homomorphic encryption, stands out for exclusively using local gradients. This increases the usability in comparison to state-of-the-art approaches and thereby opens the door to applications in control.}
\infoFootnote{J.Adamek and M. Schulze Darup are with the \href{https://rcs.mb.tu-dortmund.de/}{Control and~Cyber-physical Systems Group}, Faculty of Mechanical Engineering, TU Dortmund University, Germany. E-mails:  \href{mailto:janis.adamek@tu-dortmund.de}{\{janis.adamek}, \href{mailto:moritz.schulzedarup@tu-dortmund.de}{moritz.schulzedarup\}@tu-dortmund.de}. \vspace{0.5mm}}
\infoFootnote{\hspace{-1.5mm}$^\ast$\preprintNoteIEEEAccepted{\textcolor{red}{2024}}
}
\section{Introduction}
A common approach to improve the performance of neural network models is to increase the size of the training dataset. 
Hence, there are significant incentives for different parties to combine information from their datasets to collectively train neural networks.  
However, sharing data can be prohibited due to privacy concerns, for example in medical applications. In this case, different parties can nonetheless train a joint model by sharing the local updates of the network parameters in a federated learning (FL) scheme \cite{zhang2021survey,wen2023survey}. 	
Most common FL methods lead to models with nearby equivalent performance for all participants.
\blue{This can be} insufficient in a cross-silo setting with competitive participants \cite{huang2022cross}, as considered here. 
\blue{In fact, in such (realistic) settings, it often makes sense to award participants contributing more valuable data with more accurate models. 
This is offered by fair FL (FFL) schemes, which are in the focus of this paper.} 

The notion of collaborative fairness \cite{lyu2020collaborative} formalizes the FFL approach and it can be achieved by using different incentive mechanisms (see \cite{zhan2021survey} for a survey). In most mechanisms, the contribution of each participant is measured and aggregated in a reputation coefficient. This reputation can, e.g., be used for determining payment for participation in a monetary reward setting \cite{kang2019incentive}.
However, such a setting is not realistic in cross-silo FL, where all participants are interested in maintaining their competitive advantage. We therefore consider model-based rewards as in \cite{lyu2020collaborative}, where higher contributions lead to better performing models.

Another issue of cross-silo FL in a competitive setting is the lack of privacy against third parties. In fact, although FL has initially been invented to enhance data privacy, the central server aggregating the data (e.g., in terms of averaging weights or gradients) may infer the global model and various properties of the data distribution \cite{AonoGradients2018}. To counteract this issue, methods for privacy-preserving computations such as secure multi-party computation (SMPC) \cite{zhao2019secure}, differential privacy (DP) \cite{dwork2008differential}, or homomorphic encryption (HE) \cite{marcolla2022survey} can be applied (see Sect.~\ref{subsec:homo_enc} for an overview). 
Notably, all three methods have already been used to realize privacy-preserving classical FL (see, e.g., \cite{zhang2020batchcrypt}, \cite{kanagavelu2020two}, and \cite{wei2020federated}) \blue{and we refer to \cite{liu2022privacy} for a survey. Here, we focus on privacy of FFL schemes since we consider both privacy and fairness as crucial elements for FL in realistic and competitive setups}\orange{, e.g. for cooperating industrial companies.}

\blue{Privacy-preserving FFL (PPFFL) is still in its infancy and \blue{various open challenges have been pointed out in \cite{rafi2024fairness}}. In this context, it is remarkable that state-of-the-art FFL algorithms like \cite{xu2021gradient} do not yet include privacy measures.
Still, PPFFL has previously been realized in \cite{lyu2020towards}. However, the approach (or, more precisely, the model-based reward mechanism) builds on a intricate combination a blockchain-based architecture with procedures from DP and~HE.}

\blue{Here, in order to increase usability and expandability especially in the framework of control, we are aiming for a simplified scheme using solely HE.}
\blue{Technically,} usability is increased by eliminating the necessity of training differential privacy GANs \cite{zhang2018differentially} to create artificial data, on which the participants' contributions are measured. This tedious task is eliminated by adapting concepts from \cite{xu2021gradient} for measuring the contribution of the participants solely on its local gradient. We also expect an increased expandability to more sophisticated training methods from this step, as training differential privacy GANs for these algorithms is, to the best of the author's knowledge, not introduced in the existing literature. Furthermore, we propose a novel parameterization of the existing gradient-based reward mechanisms and observe it to be a promising new solution in our numerical experiments.  
The detailed presentation of our PPFFL scheme is organized as follows.

\textbf{Outline.} Section \ref{sec:Background} provides necessary background on FFL with model-based reward mechanisms and on HE. Our central result, i.e., a PPFFL scheme processing encrypted local gradients, is presented in Section \ref{sec:privacy_impl}.
The numerical benchmark in Section \ref{sec:Numerical_Evaluation} provides a thorough analysis of the quality and fairness of the proposed scheme. Finally, we conclude our work and discuss future research directions in Section \ref{sec:Outlook}.

\section{Background on fair federated learning and privacy-preserving computations}
\label{sec:Background}

\subsection{Fair federated learning via gradient-based rewards}
\begin{figure*}[tb]
\centering
\subfloat[ \label{subfig:Federated_Learning}]{\includegraphics[width=0.45\textwidth]{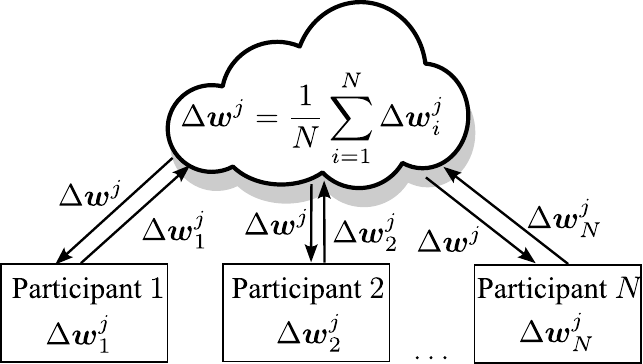}}
\subfloat[ \label{subfig:Plaintext_algorithm}]{\hspace{0.05\textwidth}\includegraphics[width=0.45\textwidth]{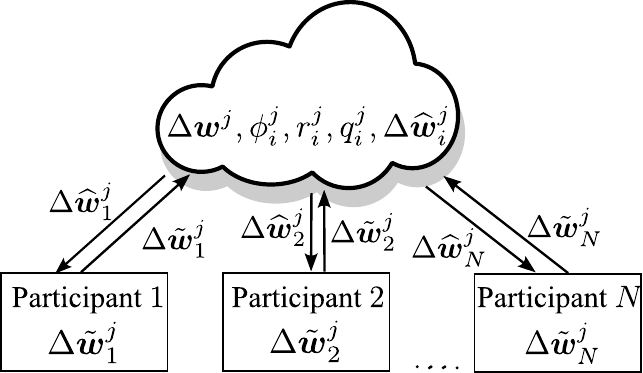}}
\caption{Illustration of (a) the FedSGD scheme and (b) the FFLX scheme, where participants send their local gradients $\Delta \wbt_i^j$ and receive reward gradients $\Delta \widehat{\wb}_i^j$ based on their reputation $r_i^j$.}
\label{fig:Plaintext_algorithms}
\vspace{-3mm}
\end{figure*}

As briefly discussed in the introduction, FL allows us to collaboratively train and improve neural network models.  
A popular realization is the FedSGD scheme \cite{mcmahan2017fedavg} depicted in Figure \ref{fig:Plaintext_algorithms}(a) for $N$ participants.  The algorithm aggregates 
 local gradients $\Delta\wb_i^j$ stemming from  local datasets in an FL gradient $\Delta \wb^j$, which is subsequently used to update all models\footnote{Here and in the following,
subscripts refer to the participants and superscripts to communication rounds.}. 
A similar approach is underlying the FFL scheme from \cite{xu2021gradient} illustrated in Figure \ref{fig:Plaintext_algorithms}(b). This scheme, to which we refer as FFLX for compactness,  will form the basis of our PPFFL scheme. Hence, we briefly summarize central steps in the following. 

The central elements of the FFLX scheme are the reputation coefficients $r_i^j$, which measure the participants' contributions to model improvements. Improvements are carried out based on the aggregated FL gradient $\Delta\wb^j$. In round $j$, this gradient is computed based on the weighted average 
\begin{equation}
\label{eq:deltaWj}    
\Delta\wb^j:=\sum_{i=1}^N r_i^{j-1} \Delta \Tilde{\wb}_i^j,  \quad \text{where}  \quad \Delta \Tilde{\wb}_i^j:=\delta \frac{\Delta \wb_i^j}{{\|\Delta \wb_i^j\|}_2}
\end{equation}
reflect normalized gradients scaled by some positive $\delta \in \R$ and where $r_i^{j-1}$ reflect reputations from the previous round. At this point, note that reputations $r_i^0$ for evaluating \eqref{eq:deltaWj} during the first round $j=1$ may be initialized based on the sizes of local datasets or simply as $1/N$.  
Now, updating the reputations is carried out as follows. First, we measure the individual contributions to the direction of the FL gradient $\Delta\wb^j$ via 
\begin{equation}
\label{eq:contributionCoeff}
\phi_i^j:=\frac{\Delta\tilde{\wb}_i^j\cdot \Delta \wb^j}{{\|\Delta\tilde{\wb}_i^j\|}_2{\| \Delta \wb^j\|}_2}.
\end{equation}
Then, we perform the initial updates $\tilde{r}_i^{j}:=\alpha r_i^{j-1}+(1-\alpha) \phi_i^j$ for some $\alpha \in (0,1)$ followed by the normalizations $r_i^j:=\tilde{r}_i^{j}/\tilde{r}^j_{\mathrm{sum}}$, where $\tilde{r}^j_{\mathrm{sum}}:=\sum_{i=1}^N \tilde{r}^j_i$.
Note that the normalizations ensure $\sum_{i=1}^N r^j_i=1$.
We then compute the relative reputations $q_i^j$ either via
\begin{equation}
\label{eq:qijVariants}
q_i^j:=\frac{\tanh (\beta r_i^j)}{\tanh (\beta r_{\text{max}}^j)} \text{ or } q_i^j=\frac{r_i^j}{ r_{\mathrm{max}}^j},  \text{ where } r_{\mathrm{max}}^j=\max_i r_i^j
\end{equation}
and where $\beta \in \R$ is a positive parameter. The second option, which does not rely on $\beta$, has been proposed in \cite{xu2020reputation} and serves as a variant for the FFLX scheme.
Once $q_i^j \in [0,1]$ has been determined, each participant gets access to the $\lfloor q_i^j \,l \rfloor$ entries with the largest absolute values from $\Delta \wb^j \in \R^l$. 
The remaining entries of the gradient $\Delta \widehat{\wb}_i^j$, which is returned to the participants, are filled up with the elements from the normalized local gradients $\Delta \tilde{\wb}_i^j$. It will turn out to be useful to formalize the procedure by introducing the function $\mathrm{mask}(\Delta\wb^j,q_i^j)$ mapping onto $\{0,1\}^l$. More precisely, the function returns a vector with entries~$1$, where the $\lfloor q_i^j \,l \rfloor$ entries with the largest absolute values appear and zeros otherwise. Using this function, we can compactly formulate the procedure as

\begin{equation}
\label{eq:deltaWijHatMask}   
\Delta \widehat{\wb}_i^j:=\mathrm{mask}(\Delta\wb^j,q_i^j) \times  (\Delta\wb^j - \Delta \tilde{\wb}_i^j) + \Delta \tilde{\wb}_i^j,
\end{equation}

where ``$\times$'' refers to element-wise multiplications of vectors. We can now infer the role of the parameter $\beta$ for the first $q_i^j$-variant in~\eqref{eq:qijVariants}. In fact, a small $\beta$ 
promotes differences between the various $q_i^j$, which may increase fairness. In contrast, a large $\beta$ increases the mean accuracy as apparent from the fact that $\beta\to\infty$ results in training with the FedSGD algorithm.

\subsection{Privacy-preserving computation via homomorphic encryption}
\label{subsec:homo_enc}

As briefly mentioned in the introduction, privacy-preserving computations can be realized using different methods. SMPC \cite{zhao2019secure} typically achieves privacy through secret sharing, where each party holds a share of private inputs and collectively, they compute a function without revealing individual inputs to each other. DP \cite{dwork2008differential} is a privacy-preserving framework for data analysis that adds controlled noise to query results to protect individual information while still providing accurate aggregated insights. HE \cite{marcolla2022survey} refers to special cryptosystems, which enable computations on encrypted data without intermediate decryption. For the desired privacy-preserving implementation of the FFLX scheme, HE seems most promising as it enables confidential computations on the central server with low communication overhead and without the need to add noise to input data (although some HE schemes use noise within the cryptosystem).

As all encryption schemes, HE schemes provide procedures to encrypt data and to decrypt the corresponding ciphertexts. We here focus on public-key schemes, where the encryption $\Enc(\xb,\pk)$ can be carried out based on a publicly available key $\pk$. For brevity, we denote the corresponding ciphertext as $\cipher{\xb}$. Its decryption via $\Dec(\cipher{\xb},\sk)$ requires the secret key $\sk$.
Now, in contrast to standard encryption schemes, HE schemes also provide procedures to carry out computations on ciphertexts. Without giving details, available operations are quite limited and typically restricted to encrypted additions and multiplications. Here, we will apply the CKKS scheme \cite{cheon2017homomorphic}, which stands out for the ability to naturally incorporate vector-valued data and operations. 
More precisely, by representing ciphertexts as elements of a ring of polynomials,  
CKKS offers procedures to realize additions $\cipher{\xb}\oplus \cipher{\yb}=\cipher{\xb+\yb}$,  element-wise  ciphertext multiplications $\cipher{\xb}\otimes \cipher{\yb}=\cipher{\xb \times \yb},$ and element-wise plaintext multiplications $\xb\cipher{\yb}=\cipher{\xb \times \yb}$ (which can also be used to multiply with scalars $a$ by means of $ a \cipher{\yb}=\cipher{a\yb}$)).  From these elementary operations, one can further derive procedures for scalar products $\cipher{\xb}\odot \cipher{\yb}=\cipher{\xb \cdot \yb}$ and subtraction $\cipher{\xb}\ominus \cipher{\yb}=\cipher{\xb-\yb}$.
These operations provide the basis for our encrypted realization of FFLX.

\section{Privacy-preserving gradient-based fair federated learning}
\label{sec:privacy_impl}

\subsection{Some notes on privacy and security} 
\label{subsec:security_models}

In the FFLX scheme, the central server has access to all gradients, including the FL gradient. Therefore, to provide a fair algorithm, a third party has to operate this server, which deteriorates the trustworthiness of the algorithm, because some data properties and the best possible update are leaked to a third party. Keeping these gradients private from the third party using HE is already a sufficient solution to all security issues related to the central server, as we model it to be semi-honest. This means the server provider may try to infer as much information as possible, but will not modify any of the calculations, which is a realistic assumption for an impartial server provider. \\
For simplicity, we will model the participants as semi-honest and address privacy issues beyond this model in subsection \ref{subsec:beyond_model}. Since honest participation in the FFLX scheme provides a fairly increased model quality, this is a realistic, albeit simplified, scenario. If all communication in the FFLX scheme is encrypted with symmetric encryptions like AES, the participants can only get access to their own local and reward gradients. Therefore, privacy against other participants can already be achieved with the FFLX scheme. We will build upon this fact and create a privacy-preserved fair federated learning algorithm for the established behavior models. 
\subsection{Details on the algorithm}
\label{subsec:privacy_algo}
\begin{figure*}[tb]
\centering
\subfloat[ \label{fig:1}]{\includegraphics[width=0.45\textwidth]{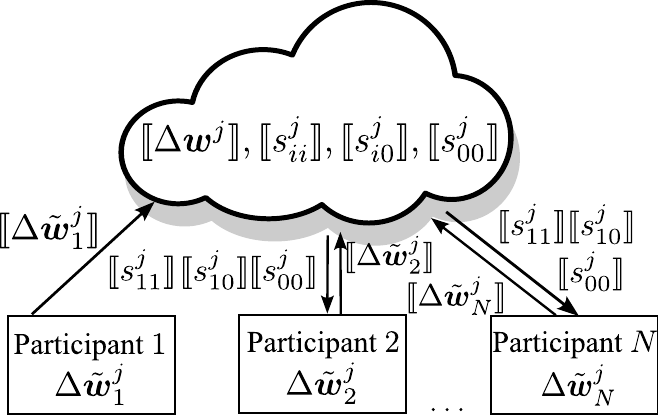}}
\subfloat[ \label{fig:2}]{\hspace{0.05\textwidth}\includegraphics[width=0.45\textwidth]{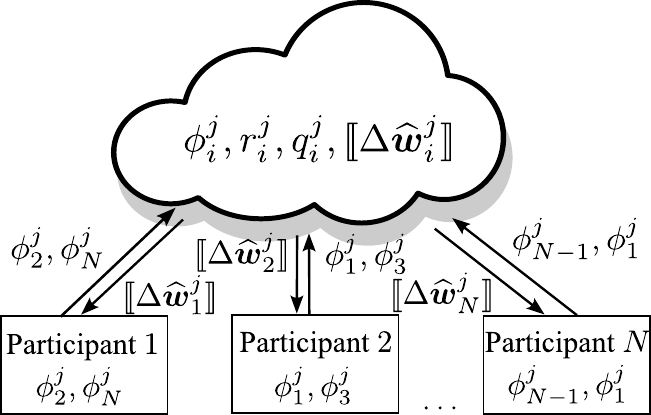}}
\caption{The two central steps of our GBPPFFL scheme (with details for participant~$1$). In (a), the calculation of the encrypted FL gradient and scalar products is depicted. In (b), the encrypted computation of the reward gradients based on the contributions $\phi_i^j$ is sketched.}

\label{fig:Encrypted_Algorithm}
\vspace{-3mm}
\end{figure*}
In this section, we will extend the FFLX scheme to the gradient-based privacy-preserving fair federated learning (GBPPFFL) algorithm, which can be accomplished using a single leveled homomorphic cryptosystem. The participants collaboratively set up this system, such that each participant has access to the secret key, whereas the central server will not be able to decrypt any ciphertexts encrypted in this homomorphic cryptosystem. All participants sharing the same secret key is no privacy threat since we will encrypt all communication with an additional symmetric encryption and share local gradients only with the central server. We have seen in subsection \ref{subsec:security_models}, that the FFLX scheme uses similar ideas, to ensure the privacy of other participants.

Analogously to the FFLX scheme, we summarize the crucial steps for the central server and the participants in Figure \ref{fig:Encrypted_Algorithm}. 
All local gradients are normalized locally, encrypted by the participants, and sent to the central server. The reputation coefficients are initialized in plaintext, as in the FFLX scheme. 
The central server then calculates the encrypted FL gradient and the scalar products 

\begin{align*}
\cipher{\Delta\wb^j}&=r_1^j\cipher{\Delta\wbt_1^j}\oplus r_2^j\cipher{\Delta\wbt_2^j}\oplus \dots \oplus r_N^j\cipher{\Delta\wbt_N^j}, \\
\cipher{s_{00}^j}&=\cipher{\Delta\wb^j}\odot\cipher{\Delta\wb^j},\\
\cipher{s_{ii}^j}&=\cipher{\Delta\wbt_i^j}\odot\cipher{\Delta\wbt_i^j} \quad \forall i\in\{1,\dots,N\},\\
\cipher{s_{i0}^j}&=\cipher{\Delta\wbt_i^j}\odot\cipher{\Delta\wb^j}\quad  \forall i\in\{1,\dots,N\}.
\end{align*} 
 We propose to send the encrypted scalar products to specific participants, who can decrypt them and calculate the contribution coefficients as in~\eqref{eq:contributionCoeff} from the decrypted scalar products in plaintext
 \begin{align}
 \label{align:new_contributions}
 \begin{split}
    \phi_i^j&=\frac{\Delta\Tilde{\wb}_i^j\cdot \Delta \wb^j}{\|\Delta\Tilde{\wb}_i^j\|_2\|\Delta \wb^j\|_2}\\&=\frac{\Delta \Tilde{\wb}_i^j\cdot \Delta \wb^j}{
    \sqrt{\Delta \Tilde{\wb}_i^j\cdot \Delta \Tilde{\wb}_i^j}\sqrt{\Delta \wb^j\cdot \Delta \wb^j}
    }
    =\frac{s_{i0}^j}{\sqrt{s_{ii}^j s_{00}^j}}.
 \end{split}
\end{align}
As the participants cannot infer the underlying gradients from the scalar products, this leads to a less complex calculation, since the calculation of square roots and the division are very demanding with state-of-the-art homomorphic cryptosystems. We opt for redundantly calculating $\phi_i^j$ at the direct neighbors of participant $i$, considering a circular setting, as it is shown in Figure \ref{fig:Encrypted_Algorithm}. This approach is valid for the semi-honest model and can address further security issues, which we will discuss in subsection \ref{subsec:beyond_model}.  
Subsequently, the server updates the $r_i^j$ equivalently to the FFLX scheme, with the uploaded contributions from \eqref{align:new_contributions} and calculates the $q_i^j$ according to \eqref{eq:qijVariants}. 
In our GBPPFFL scheme, we also consider an alternative calculation of the relative reputation coefficients, which can lead to a better separation of datasets with different quality. We update the $r_i^j$ according to the FFLX scheme, again with the uploaded $\phi_i^j$ from \eqref{align:new_contributions}, use the parameter-free calculation of $q_i^j$ from \eqref{eq:qijVariants} and then apply the modification
\begin{align}
\label{align:new_normalization}
    q_i^j\leftarrow(q_i^j)^{\frac{1}{\gamma}}
\end{align} for some positive $\gamma\in\R$, where ``$\leftarrow$'' stands for an assignment. As a result $q_i^j$ stays in the interval $[0,1]$, but the number of retained elements will be more distinguished with a smaller value of $\gamma$. Therefore, it serves the same purpose as $\beta$ and should also increase the mean accuracy with higher values, as we are obviously obtaining the FedSGD algorithm for $\gamma\to \infty$. Note, that we can make arbitrary changes to these calculations, as they are executed in plaintext. The plaintext calculation is also used by \cite{lyu2020towards} and in accordance with our privacy goal since the central server cannot infer any gradients from the contributions or reputation coefficients.

Due to the CKKS ciphertext encoding, we cannot trivially access single entries of the gradient vectors. Therefore, the reward gradients are calculated in an encrypted fashion by multiplying with a plaintext mask that contains the elements
\begin{align*} 
    \mb_i^j(k)=\begin{cases}
        1&\text{if the $k$-th element }  \text{in round } j \\&\text{is retained for participant } i,\\
        0&\text{otherwise.}
    \end{cases}
\end{align*}
Similar to the FFLX scheme, we retain $\lfloor q_i^j l\rfloor$ elements of the FL gradient. To avoid complex sorting algorithms, we define a randomized sequence of elements that are retained first, which is sampled for every gradient calculation. Since this is a different heuristic than using the entries with the largest absolute values, we will investigate the effects on the fairness of the algorithm in our numerical section. However, we emphasize that, due to the normalization, the participant with the highest data quality will obtain the same amount of FL gradient entries and one only sees a difference in the spread of the solution quality. As the parameters $\beta$ and $\gamma$ offer the opportunity to directly influence this spread, we expect only an adaption of these parameters to obtain similar results. After the masking, the local gradient is added at the zero entries, which can be formulated as $\cipher{\Delta\hat{\wb}_i^j}=\mb_i^j {(\cipher{\Delta\wb^j}\ominus\cipher{\Delta\Tilde{\wb}_i^j})}\oplus\cipher{\Delta\Tilde{\wb}_i^j}$ analogously to~\eqref{eq:deltaWijHatMask}.
Only masking and locally replacing the zeros with the local gradients is a valid alternative, but requires additional communication overhead or zero detection algorithms, because either the ordered set of indices and the $q_i^j$ would have to be communicated, or the zeros have to be detected by the local participants.

\subsection{Security issues beyond our model}
\label{subsec:beyond_model}
\textbf{Modifications.} A trivial security issue arises from involving the participants in the calculation of the contributions, since the participants will, if not modeled as semi-honest, try to increase their own and decrease other participants' contributions. We have already presented one possible solution, in which two participants other than participant $i$ redundantly calculate $\phi_i^j$. The server can check the correctness of the result, which prevents malicious activities, as long as there is no collaboration between the two participants. Although this is a realistic scenario in a competitive setting, especially because participant $i$ is not involved, we can also increase security by introducing more redundancy.

\textbf{Collaboration.} As the central server is modeled as impartial and semi-honest, we rule out collaborations between a participant and the central server. However, it is obvious, that the current scheme is not privacy-preserving if such a collaboration would be initiated, since this group has access to all gradients and the secret key. 
One way to eliminate this issue is to use multi-key HE techniques like those described by \cite{ma2022privacy}. Remarkably, the HE also allows to eliminate the central server, as an aggregation could take place at one of the participants if this participant does not have the secret key. As collaborations between the participants are much less likely in a competitive setting, the correctness of the calculation can again be ensured by independently calculating the reward gradients multiple times and comparing the obtained results. Related ideas can be reviewed in the contribution of \cite{lyu2020towards}. Although excluding the possibility of collaboration seems to be straightforward, it is beyond the scope of this contribution. 
\section{Numerical Evaluation}
\label{sec:Numerical_Evaluation}
\subsection{Experimental Setup}
Although our long-term goals are applications from the field of control theory, where a gradient-based reward mechanism may be more advantageous, we will consider a similar experimental setup as used by \cite{xu2021gradient} to compare our GBPPFFL scheme with a state-of-the-art gradient-based reward mechanism. The algorithms are tested on making appropriate decisions on the image classification datasets MNist \cite{lecun-mnisthandwrittendigit-2010} and Cifar10 \cite{krizhevsky2009learning} as well as the text classification datasets movie review (MR) \cite{Pang+Lee:05a} and Stanford sentiment treebank (SST)~\cite{socher2013recursive}.

As data distributions, we will consider both the identically and independently distributed data (IID) and a non-IID (NIID) setting. The IID case is created by randomly distributing a subset of the dataset onto $N$ participants. However, to test the fairness of the FL algorithm, we will further consider a uniformly as well as a non-uniformly distributed number of data samples between the participants. The uniform case leads to local dataset sizes of $\{600,2000,1919,1709\}$ for MNist, Cifar10, MR and SST dataset. The second case is generated by using a power law relationship between the individual dataset sizes, as it is described by \cite{xu2021gradient}. Each participant therefore gets a different dataset size, exemplary ranging from $71$ to $1120$ elements for the MNist dataset, where we consider $10$ participants.

Even if the underlying distribution of the data should be equivalent, we will investigate whether the quality advantage due to a higher amount of accessible data is preserved using our GBPPFFL scheme.
As a second option to diversify the quality of the local datasets, the NIID case is only created for image classification datasets and constructed by inserting a different number of available classes in the local datasets (e.g., numbers in the MNist dataset).  
For an appropriate measurement of the quality difference, an independent and large test dataset is provided (e.g., $10000$ images of the MNist dataset), which contains all classes.

Our numerical evaluation involves three steps. First, we will consider a case study on the different parameterized reward strategies for the GBPPFFL scheme and afterward compare our encrypted solution with a plaintext simulation of the algorithm. After that, we provide a thorough evaluation on the aforementioned datasets and data distributions. On the MNist dataset, we will also investigate the influence of the number of participants, using $N\in \{10,20\}$, whereas the number of participants will be $N=10$ for the Cifar10 dataset and $N=5$ for both text datasets. As neural network models, we use typical combinations of CNN and fully connected layers for the image datasets and train them using classical stochastic gradients.
 
The text datasets are approached by text embedding CNN models, which are trained with the Adam optimizer \cite{kingma2014adam}. In all our experiments, we set $\alpha=0.95$ and $\delta$ to $0.5$ for the MNist dataset, to $0.15$ for Cifar10, and to $1$ for both text classifications.

While the accuracy on the test dataset serves as the measure of the quality of the solution, we measure collaborative fairness with the Pearson coefficient between standalone accuracy and FL accuracy for all participants. This procedure is introduced by \cite{lyu2020collaborative}, with the Pearson coefficient of two vectors $\xb$ and $\yb$ being defined as ${\rho=\frac{\sum_i (\xb_i-\bar{\xb})(\yb_i-\bar{\yb})}{\sqrt{\sum_i (\xb_i-\bar{\xb})^2\sum_i (\yb_i-\bar{\yb})^2}}.}$ In our case the vector entry $x_i$ equivalents the resulting classification accuracy on the test dataset, when participant $i$ trains solemnly on his own dataset and $y_i$ represents the same accuracy when using the FL scheme. Subsequently, we will denote the training on the local datasets as standalone method in all our experiments. 
\subsection{Case study on the parameterized reward mechanism}
We will not fine-tune the parameters $\beta$ or $\gamma$ for our numerical analysis on all datasets and configurations in Subsection \ref{subsec:full_results}, but use the parameter-free variant from~\eqref{eq:qijVariants}. 

However, in this case study we investigate the basic feasibility of adjusting the fairness of the GBPPFFL scheme for the MNist dataset with $N=10$. In Table \ref{table:results_case_study}, the accuracies and the Pearson correlation coefficients for the non-uniform IID case and the NIID case are displayed for different parameters. The first columns refer to the GBPPFFL scheme with the parameterized calculation of $q_i^j$ from \eqref{eq:qijVariants}. There we cannot reproduce the trend of increasing mean accuracy and decreasing fairness for higher values of $\beta$. The last columns address our proposed solution from \eqref{align:new_normalization}, which seems to be working as expected on the NIID case, this time decreasing the fairness and increasing the mean accuracy with larger values of $\gamma$. Although the fairness remains nearby constant, our solution can also distinguish the mean accuracies in the non-uniform IID setting in the expected way. Therefore, our parameterization method seems to be a valuable extension to existing work and should be further investigated for finding a suitable compromise between fairness and mean accuracy in PPFFL algorithms. 
\begin{table}[tb]
	\centering
	\begin{tabular}{cccccc}
 \toprule
		&$\beta$/$\gamma$& \multicolumn{2}{c}{GBPPFFL ($\beta$)}& \multicolumn{2}{c}{GBPPFFL ($\gamma$)} \\
\cmidrule(lr){3-4}\cmidrule(lr){5-6}
		
		&Data Split&IIDNU&NIID&IIDNU&NIID\\
  		\midrule
    \parbox[t]{4mm}{\multirow{5}{*}{\rotatebox[origin=c]{90}{Accuracies}}}
		&Standalone&${90}({94})$&${53}({93})$&${90}({94})$&${53}({93})$\\
		&$0.5$/$0.1$&${96}({97})$&${63}({95})$&$92(96)$&$54(93)$\\
		&$1.0$/$0.2$&${96}({97})$&$57(94)$&$93(96)$&$56(93)$\\
		&$1.5$/$0.5$&${96}({97})$&$62({95})$&$94(96)$&$62({95})$\\
		&$2.0$/$1$&${96}({{97}})$&${63}({95})$&${96}({97})$&${62}({95})$\\
		\midrule
		\multirow{4}{*}{$\rho$}
		&$0.5$/$0.1$&${1.00}$&${0.94}$&${1.00}$&${1.00}$\\
		&$1.0$/$0.2$&${1.00}$&${0.95}$&${1.00}$&$0.98$\\
		&$1.5$/$0.5$&${0.99}$&${0.95}$&${0.99}$&$0.98$\\
		&$2.0$/$1$&${1.00}$&${0.94}$&${1.00}$&${0.94}$\\
		\bottomrule
	\end{tabular}
	\caption{Mean and maximum accuracies and fairness coefficients for the case study using the parameterized mechanism from \eqref{eq:qijVariants} for the first columns and \eqref{align:new_normalization} for the last columns. Maximum accuracies are in parentheses. IIDNU: Non-uniformly distributed IID case}
	\label{table:results_case_study}
  
\end{table}

\setlength{\tabcolsep}{4pt}
\begin{table*}[tb]
\centering
    \begin{tabular}{ccccccccccccccc}
    \toprule
    &&\multicolumn{6}{c}{MNist}&\multicolumn{3}{c}{Cifar10}&\multicolumn{2}{c}{MR}&\multicolumn{2}{c}{SST}\\
    \cmidrule(lr){3-8} \cmidrule(lr){9-11}\cmidrule(lr){12-13}\cmidrule(lr){14-15}
    \multicolumn{2}{c}{$N$}&\multicolumn{3}{c}{$10$}&\multicolumn{3}{c}{$20$}&\multicolumn{3}{c}{$10$}&\multicolumn{2}{c}{$5$}&\multicolumn{2}{c}{$5$}\\
    \midrule
    \parbox[t]{4mm}{\multirow{5}{*}{\rotatebox[origin=c]{90}{Accuracies}}}
    &Data Split&IIDU&IIDNU&NIID&IIDU&IIDNU&NIID&IIDU&IIDNU&NIID&IIDU&IIDNU&IIDU&IIDNU\\
    \cmidrule(lr){3-5} \cmidrule(lr){6-8}  \cmidrule(lr){9-11}\cmidrule(lr){12-13}\cmidrule(lr){14-15}
    &Standalone &${93}({94})$&${91}({95})$&${53}({93})$&${92}({94})$&${91}({94})$&${49}({93})$&${35}({39})$&${33}({38})$&${25}({36})$&${56}({60})$&${53}({60})$&${33}({34})$&${30}({33})$\\
    &FFLX&${97}({97})$&${97}({97})$&${67}({95})$&${97}({98})$&${97}({98})$&${60}({95})$&${44}({45})$&${43}({45})$&${28}({43})$&${82}({82})$&${70}({82})$&${38}({38})$&${33}({36})$\\
    &GBPPFFL  &${97}({97})$&$96({97})$&$63({95})$&${97}({98})$&${97}(97)$&$56({95})$&$38({39})$&$39(41)$&${28}(42)$&${80}({80})$&${67}({78})$&${39}({39})$&${33}({36})$\\
    \midrule
    \multirow{2}{*}{$\rho$}
    &FFLX&${0.47}$&${0.99}$&${0.92}$&${0.37}$&${0.99}$&${0.94}$&${0.16}$&${0.84}$&${0.89}$&${0.47}$&${0.96}$&${0.22}$&${0.96}$\\
    &GBPPFFL  &${0.59}$&${1.00}$&${0.94}$&${0.37}$&${0.97}$&${0.96}$&${0.46}$&${0.83}$&${0.91}$&${0.38}$&${0.84}$&${0.26}$&${0.98}$\\
    \bottomrule
    \end{tabular}
    \caption{Mean and maximum (in parentheses) accuracies (in $\%$) and fairness coefficients for all datasets. 
    IIDU: Uniformly distributed IID case, IIDNU: Non-uniformly distributed IID case.}
    \label{table:results_all}
    \vspace{-3mm}
\end{table*}
\subsection{Case study on the encrypted implementation}

\textbf{Cryptosystem.} As mentioned in Section \ref{sec:privacy_impl}, we opt for the CKKS \cite{cheon2017homomorphic} cryptosystem, which is a state-of-the-art leveled homomorphic cryptosystem based on the learning with errors problem over polynomial rings (RLWE). In our implementation, we use the OpenFHE \cite{al2022openfhe} library and Pybind11 \cite{pybind11} for the Python integration. As parameters we chose a ring dimension of $N_r=2^{14}$, a scaling factor of $\Delta=2^{50}$ and a first modulus of $Q=2^{60}$. This provides a sufficient precision for our multiplicative depth of $2$ and is coherent with a $128$ bit security as described in the homomorphic encryption standard \cite{albrecht2021homomorphic}. We can further assume that a $128$ bit security setting is sufficient for keeping encrypted data private since that is a well-established security standard.

\textbf{Computational complexity.} For a better comparison, we use equivalent network architectures as \cite{xu2021gradient}, which leads to about $10^5$ network parameters for the MNist dataset. Because only vectors of a maximal length of $\frac{Q}{2}$ can be stored in one ciphertext, we split the calculations into different instances, since a large ring dimension is impractical. Both the encrypted scalar product and the multiplication with a mask can be trivially decomposed to multiple ciphertexts because they are linear operations. As a result, the complexity of the FL scheme largely depends on the number of network parameters, but many computations are parallelizable. 
We used an AMD Ryzen 7 5800H processor with $8$ GB RAM and an NVIDIA GeForce RTX 3050 Ti GPU with $4$ GB RAM for the training and encrypted FL scheme. With this limited hardware and without parallelization, we achieve a computation time per round of roughly $50\,\mathrm{s}$ with and $0.1\,\mathrm{s}$ without the encryption, not considering the communication. Parallelization up to $20$ processes should be trivial due to the splitting of the calculation and the huge computational power of the central server. The resulting overall computation time is therefore still reasonable, so the privacy preservation can be realistically reached.\\
\textbf{Comparison.} Due to the increased computational overhead of using the encryption, we will evaluate the GBPPFFL scheme with a simulated encryption. We can show, that even the effect of differences in the stochastic gradient are comparable to those of the low approximation error. As an example, the non-uniform IID case on the MNist dataset with $N=10$ is evaluated like in table \ref{table:results_all}. A comparable mean accuracy of $95\%$, maximum accuracy of $97\%$, and fairness coefficient of $0.98$ is achieved with the encrypted algorithm. We can also reproduce the $97\%$ accuracy in the uniform IID case. Because we typically observe up to $1\%$ accuracy difference between two optimization runs for these cases, we can assume, that our simulation of the encrypted algorithm is a valid approximation.

\subsection{Full experimental results}
\label{subsec:full_results}

The mean and maximum achieved accuracies of all image and text datasets and configurations on the test dataset can be found in Table \ref{table:results_all}. As we could not entirely reproduce the results from \cite{xu2021gradient}, but had access to their implementation, we re-evaluated the procedure for a better comparison on our hardware, using the non-parameterized variant from \eqref{eq:qijVariants}.

We can show, that the accuracy of the model of every participant is increased by the FFLX and GBPPFFL scheme in both the IID and NIID case in comparison to the standalone performance. Due to the choice of the most simple reward mechanism, the slightly lower mean accuracy of the GBPPFFL scheme is expected, as we do not retain the most important gradients, but a random subset. The maximum accuracy is comparable, as both algorithms retain the complete reward gradient. The main objective of this research, which is to safeguard the gradients' privacy to third parties without a significant loss in performance of the application, is therefore achieved.

A difference in the quality of the local data is maintained in the resulting accuracy by both algorithms. In both the non-uniform IID case and the NIID case we achieve correlation coefficients above $0.83$ on all image and text datasets. This is a strong correlation and the fairness is particularly high on the MNist dataset. A loss of fairness in the GBPPFFL scheme is not observed, so the GBPPFFL heuristic for a fair reward mechanism seems to be satisfactory. In some cases, the fairness can even be increased, for example for $10$ participants on the MNist dataset. These results seem to be robust regarding the number of participants, which is an important feature for the considered use case.\\
The Pearson correlation coefficient in the uniformly distributed IID scenario is relatively low. This is due to the fact, that the quality difference in the standalone training phase is already small and most likely influenced by randomly distributed differences in the training process. The small difference in performance is preserved by both algorithms and due to these influences, different participants than before get slightly better models, which is already an issue in the FFLX scheme. Considering our setup this is not a significant disadvantage, since the idea of this configuration is that all participants hold data of equal quality. Therefore, a classical FL algorithm would be already sufficient, which leads to similar correlation coefficients, according to \cite{xu2020reputation}.

\section{Conclusion and Outlook}
\label{sec:Outlook}
We introduced, to the best of our knowledge, the first privacy-preserving and collaborative fair model-based reward mechanism for FL, which is exclusively depending on the local gradients. 
This GBPPFFL scheme can achieve comparable accuracies and fairness as state-of-the-art gradient-based reward mechanisms without privacy protection against third parties on image and text classification datasets. Furthermore, we can show that the computational overhead is still appropriate and we can adjust the fairness of the algorithm using parameterized reward mechanisms, to which we contribute a novel and promising variant.

Future research directions are threefold. First, we acknowledge that some information is leaked by calculating the reputation coefficients in plaintext. To address this issue, a completely encrypted implementation is a central aim for the future, e.g., by using SMPC.
Second, SMPC and multi-key HE are also promising techniques for providing privacy guarantees under collaboration, which we will also address in subsequent work.
Third, we will explore the advantage of exclusively gradient-based reward mechanisms for the privacy preservation of more sophisticated training algorithms to which FL can be applied. Such examples include federated reinforcement learning or imitation learning for control applications.

\bibliographystyle{IEEEtran}

\end{document}